\documentclass{article}
\usepackage{spconf,amsmath,graphicx}

\usepackage{cite}
\usepackage{amsmath,amssymb,amsfonts}
\usepackage{algorithmic}
\usepackage{graphicx,color}
\usepackage{textcomp}
\usepackage{bbm}
\usepackage{bm}
\usepackage{caption}
\usepackage{graphicx, subfig}
\usepackage {booktabs}
\usepackage {bm}
\usepackage[colorlinks,
            linkcolor=blue,
            anchorcolor=blue,
            citecolor=blue]{hyperref}
\usepackage{multirow}

\title{L-SNet: From region localization to scale invariant medical image segmentation } 
\name{Jiahao Xie, Sheng Zhang, Jianwei Lu$^{\star}$, Ye Luo$^{\star}$}
  
\address{ School of Software Engineering, Tongji University, China }
\begin{document}
%
\maketitle
\begin{abstract}
\vspace{-3pt}
Coarse-to-fine models and cascade segmentation architectures are widely adopted to solve the problem of large scale variations in medical image segmentation. 
However, those methods have two primary limitations: the first-stage segmentation becomes a performance bottleneck; the lack of overall differentiability makes the training process of two stages asynchronous and inconsistent. In this paper, we propose a differentiable two-stage network architecture to tackle these problems. In the first stage, a localization network (L-Net) locates Regions of Interest (RoIs) in a detection fashion; in the second stage, a segmentation network (S-Net) performs fine segmentation on the recalibrated RoIs; a RoI recalibration module between L-Net and S-Net 
eliminating the inconsistencies.
Experimental results on the public dataset show that our method outperforms state-of-the-art coarse-to-fine models with negligible computation overheads. 
\end{abstract}
%

%
%

\vspace{-12pt}
\section{Introduction}
\label{sec:introduction}
\vspace{-10pt}

Although various models have made significant achievements on the automated medical image segmentation \cite{unet,FCN,segnet,psp}, the important issue on scale variations remains unsolved, which is more salient in medical image processing since scales of different organs or lesions often vary prodigiously. Besides, proportions of the foreground objects on a whole image also vary greatly.
Currently, various research has been devoted to solving or mitigating these problems on the pancreas CT dataset (Pancreas-CT, available on TCIA\cite{TCIA}), on which the scale variation problem is especially obvious. \par

\begin{figure}[t]\vspace{-20pt}
    \centering
    \resizebox{0.45\textwidth}{!}{
    \includegraphics{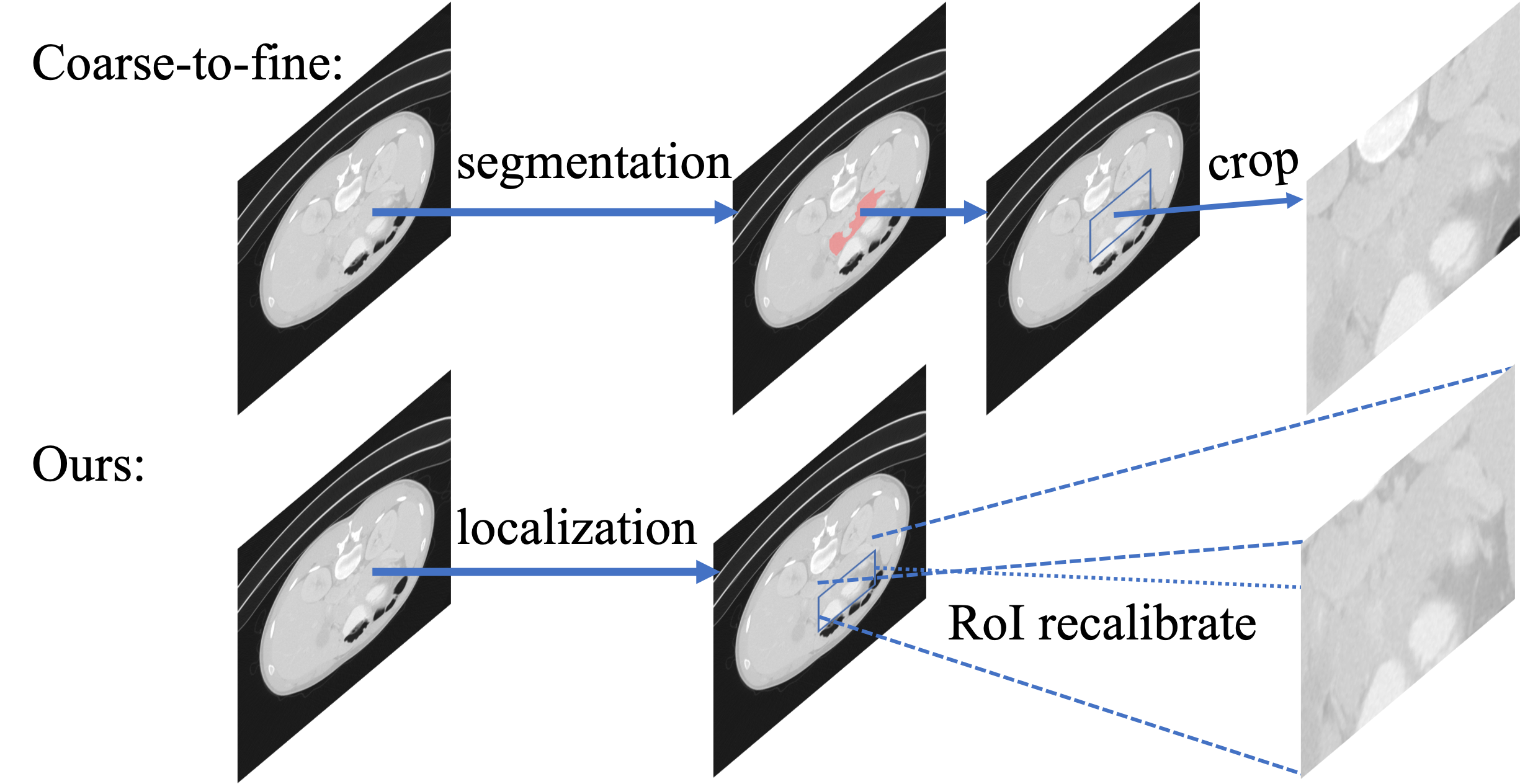}
    }\vspace{-10pt}
    \caption{The main difference between the widely used coarse-to-fine method and our L-SNet, before sending RoIs to the second stage for fine segmentation.}
    \label{fig:intro}\vspace{-15pt}
\end{figure}

Recent researchers designed two-stage cascade models \cite{zhou2017fixed,3dc2f} to mitigate the scale imbalance in a coarse-to-fine way where the first model conducts coarse segmentation and roughly locate Region of Interests (RoIs), and the second model performs finer segmentation on the located RoIs. Although this approach has achieved state-of-the-art performance on the pancreas dataset, it suffers from a few limitations: inconsistencies between training and testing, difficulties in tuning, etc. 
Though RSTN \cite{RSTN}, which is inspired by and improved the coarse-to-fine model with a saliency transformation module, can jointly optimize two deep networks, its optimization of locating RoIs is indirect and not interpretable. Further, the recurrent branch incurs high computational complexities. 
Attention U-Net \cite{attentionunet} alleviates the problem brought by large scale variations through the proposed attention gates, which merge both low-level and high-level features based on the global context of each. However, attention U-Net achieves a low performance gain. Unlike all of them, our entirely-differentiable architecture L-SNet solves large scale variations and eliminates many limitations of the coarse-to-fine models. Fig.\ref{fig:intro} shows our improvements compare to coarse-to-fine method. \par

\begin{figure*}[ht]\vspace{-10pt}
    \centering
    \resizebox{\textwidth}{!}{
    \includegraphics{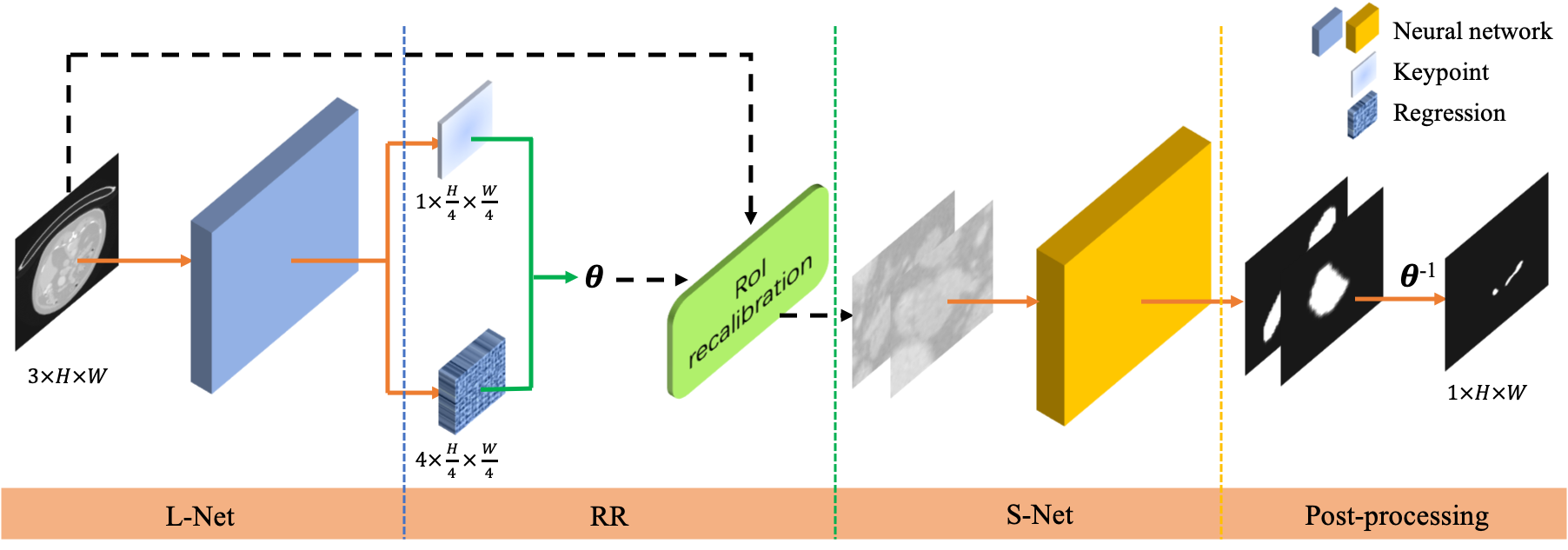}
    }\vspace{-10pt}
    \caption{The overview of our proposed L-SNet. Given a CT image slice, L-Net is to locate RoIs, and the RoI recalibration module(RR) is to recalibrate located RoIs into a fixed scale for finer segmentation by S-Net. The post-processing is to restore finely segmented masks to their original shapes and positions on the input image. Our flexible architecture poses few restrictions on the form of L-Net and S-Net since they can be any CNNs.}
    \label{fig:process}\vspace{-10pt}
\end{figure*}

In this paper, we proposed an innovative two-stage network architecture, L-SNet, which uses L-Net to resolve the first location problem and use S-Net to perform the fine segmentation. As is shown in Fig.\ref{fig:process}, in the first stage, L-Net predicts and locates all RoIs; in the second stage, S-Net conducts finer segmentation; RoI recalibration module recalibrates RoIs, bridging two stages and making L-SNet entirely-differentiable. 
The main contributions of our work can be summarized as follows:
(1) We propose an innovative two-stage network architecture to solve the large scale variations, where the first stage conducts effective RoI detection rather than widely adopted coarse segmentation.
(2) We design an interpretable RoI recalibration module to bridge the gradients propagation between L-Net and S-Net, making L-SNet entirely differentiable. 
(3) Our proposed L-SNet consistently improves the coarse-to-fine models' performance on the Pancreas-CT dataset with measly computation overheads.

\vspace{-15pt}
\section{The proposed methods}
\label{sec:the proposed methods}
\vspace{-10pt}
The overall architecture of our proposed method is displayed in Fig.\ref{fig:process}, which is mainly composed of three components: the localization Network (L-Net), the RoI Recalibration (RR) module, and the Segmentation Network (S-Net). 
Details of each component are introduced in the following sub-sections. \par

\vspace{-10pt}
\subsection{Localization Network}
\vspace{-5pt}

L-Net locates each RoI using a bounding box (bbox) by predicting the location of its center point and the distances from that point to four boundaries of the box, which involves predicting six parameters---two for the center point's location $(x,y)$ and four for the distances $(l,r,t,b)$ from the center point to the left, right, top, and bottom edge of the bbox. This process can be decomposed into two tasks: keypoints prediction and distance regression. Therefore, the RoI can be accurately located by the bbox with a topleft anchor $(x-l,y+t)$ and a bottom right anchor $(x+r,y-b)$. \par



\begin{figure}[t]\vspace{-5pt}
    \centering

    \resizebox{0.45\textwidth}{!}{
    \includegraphics{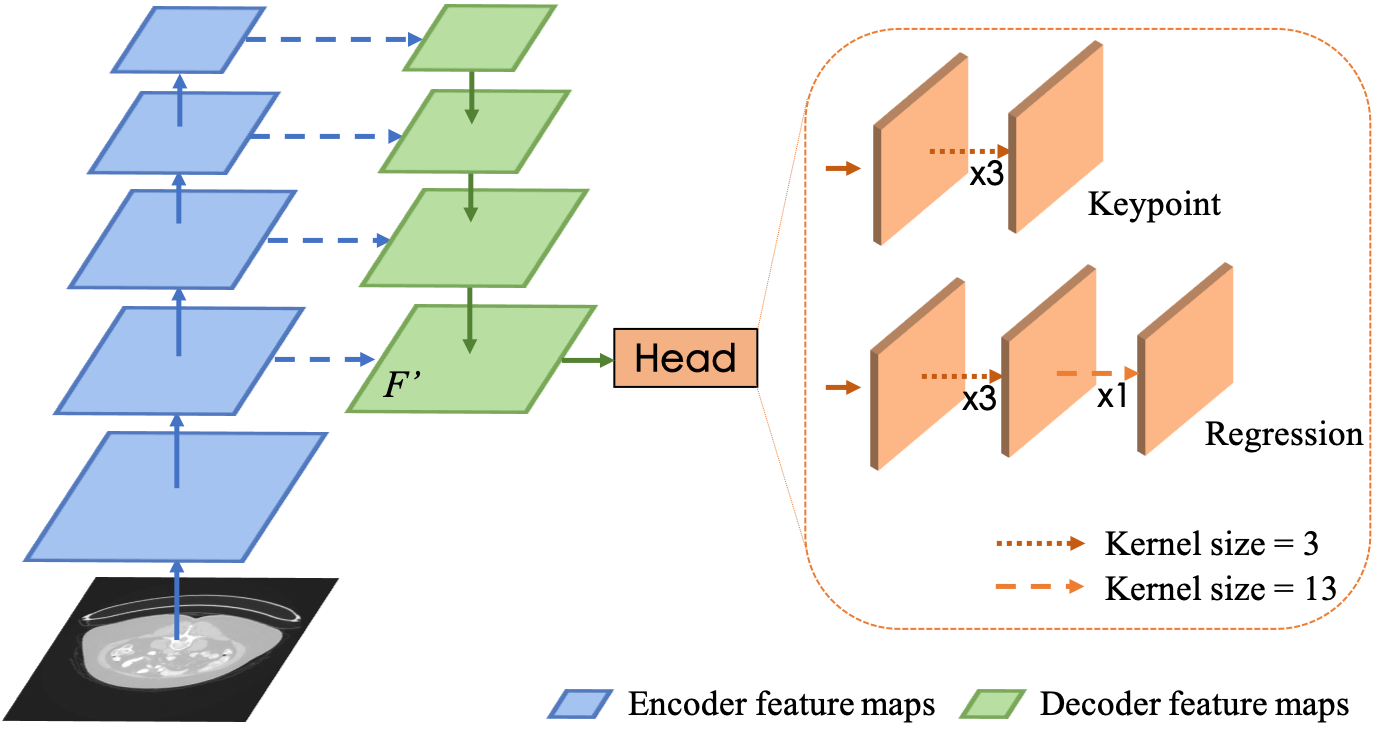}
    }\vspace{-10pt}
        \caption{The pipeline of L-Net. Feature map $F^{\prime}$ is extracted by a encoder-decoder structure. Two branches perform keypoints prediction and bounding box regression on $F^{\prime}$, respectively.}
    \label{fig:L-Net}\vspace{-15pt}
\end{figure}

\textbf{Network architecture.}
Aiming to approximate the locations of foreground objects, we design the structure of L-Net as an encoder-decoder as most models do \cite{unet, deeplab, hourglass}. In practice, it can flexibly be any reasonable basic structure, such as FCN\cite{FCN} and U-Net\cite{unet}, with little modifications. \par

L-Net takes images with the size of $H \times W$ as input, and their RoIs are predicted and located in an anchor-free fashion by outputs of two branches: the keypoints prediction (KP) head, for locating each keypoint's position $(x,y)$, and the bounding box regression (BBR) head, for predicting distances $(l,r,t,b)$ for each keypoint. \par 

Taking the basic structure U-Net as an example (see Fig.\ref{fig:L-Net}), given the input, it is first sequentially down-sampled five times by the encoder to obtain a feature map with a size of $\frac{H}{32} \times \frac{W}{32}$, which is then upsampled three times by the decoder into a feature map $F^{\prime}$ with a size of $\frac{H}{4} \times \frac{W}{4}$. A center-aligned sampling strategy is adopted on the feature map $F^{\prime}$ so that each position $(x,y)$ can be mapped back to the position $(4x+2, 4y+2)$ on the input image.
\par

\textbf{Keypoints prediction.}
\label{subsec:keypoints-prediction} 
An ideal keypoint of the foreground object is its center point. Instead of computing the coordinates of the keypoints directly, in this paper, our KP head predicts keypoints by outputing a heatmap $ \bm{K} \in \mathbb{R}^{1 \times \frac{H}{4} \times \frac{W}{4}}$ to approximate the ground-truth heatmap $ \bm{K}^{*} \in \mathbb{R}^{1 \times \frac{H}{4} \times \frac{W}{4}}$ which is generated by the true centers of the foreground bboxes. In details, we use the ground-truth heatmap to guide another heatmap generation, and then locate keypoints on the generated heatmap by thresholding (Details in Section \ref{subsec:training-and-inference}).
Here, each element $K_{x,y}$ of $\bm{K}$ represents the output probability of the point $(x,y)$ being a foreground keypoint. 
Each element $K^{*}_{x,y}$ of $ \bm{K}^*$, similar to \cite{duan2019centernet,cornernet}, is defined as:
\begin{align}
    \vspace{-5pt}
    K^*_{x,y} &= \min{(1, \sum_{(x_c,y_c) \in S}{e^{-\frac{d_x^2 + d_y^2}{2 \delta}}})}
    \vspace{-5pt}
\end{align}
\noindent where $d_x= \max{(0, |x - x_c|-\beta)}$, $d_y= \max{(0, |y - y_c|-\beta)}$. %
$S$ denotes the set of the center points of all foreground objects on $ \bm{K}^*$, and $(x_c, y_c)$ is the location of a center point in $S$. 
${K}^{*}_{x,y}$ represents  the probability of point $(x,y)$ to be a true keypoint. 
Here, the heatmap is generated in the form of Gaussian function with the variance $\delta$. $\beta$ controls its tolerant biases. We set $\delta=20$ by default and $\beta=2$ empirically. \par

\textbf{Bounding box regression.}
\label{subsec:bounding-box-regression}
Once the predicted key points are obtained, the BBR head predicts the bbox at the key point. Specifically, the BBR head regresses the $ \bm{T} \in \mathbb{R}^{4 \times \frac{H}{4} \times \frac{W}{4}}$ to approximate the ground-truth $\bm{T}^* \in \mathbb{R}^{4 \times \frac{H}{4} \times \frac{W}{4}}$. Here, each element ${T}_{x,y}$ of $\bm{T}$ represents the predicted values of $(l,r,t,b)$ at the point $(x,y)$. Each $ {T}^*_{x,y}$ on $ \bm{T}^*$ is a 4D vertor $(l^*,r^*,t^*,b^*)$ denotes the ground-truth point-edges distances of position $(x,y)$. 
Later, we will show in our loss function that only the points located around the center of ground-truth bbox region have values of $ {T}^*_{x,y}$. We predict $ \bm{T}$ by two convolutions with kernel size 3 followed by one convolution with kernel size 13 based on the feature map $F^{\prime}$. The convolution with a larger kernel has larger receptive fields, which is more beneficial to the distance regression task. \par




\vspace{-12pt}
\subsection{RoI recalibration}
\vspace{-5pt}

After RoIs with variant sizes are located, in order to feed S-Net with uniformized shapes of RoIs, a differentiable RoI Recalibration module (RR) is designed. Inspired by the grid generator in STN \cite{STN}, we design RR to recalibrate the located RoI by mapping each point $(x,y)$ on the RoI to a new position $(\Acute{x},\Acute{y})$ through an affine transformation which is defined as:
\vspace{-5pt}

\begin{equation}
    \left[
    \begin{matrix}
    \Acute{x} \\ \Acute{y}
    \end{matrix}
    \right]
     = 
    \theta
    \left[
    \begin{matrix}
    x \\ y \\ 1
    \end{matrix}
    \right]
    =
\left[
    \begin{matrix}
    s_x & 0 & b_x \\
    0 & s_y & b_y 
    \end{matrix}
\right]
\left[
\begin{matrix}
x \\ y \\ 1
\end{matrix}
\right]
\end{equation}
where $\theta$ is the affine transformation matrix with parameters $\{s_x, s_y, b_x, b_y\}$. Here, $s_x$ and $s_y$ denotes scaling factors of $x$-axis and $y$-axis, and $b_x$ and $b_y$ are horizontal and vertical translations. 
In order to back-propagate gradients, the four parameters $\{s_x, s_y, b_x, b_y\}$ are converted from the L-Net outputs (i.e., a keypoint $(x,y)$ and its associated bbox regression result $(l,r,t,b)$) as:
\vspace{-5pt}
\begin{align}
&s_x = \frac{l + r + \alpha}{W}, &b_x = x + \frac{r - l}{2} - \frac{W}{2}, \notag\\
&s_y = \frac{t + b + \alpha}{H}, &b_y = y + \frac{b - t}{2} - \frac{H}{2},
\label{f:totetha}\vspace{-5pt}
\end{align}
where $\alpha$ is the border margin of extra pixels padded around the localized RoIs, which is set as 15 in this paper.

It is worthy of remarking RR module's superiorities: coordinates are sampled and predicted in Float type in L-Net, and RR module sampled image by bilinear interpolation rather than hard cropping. So RR module eliminates the misalignment caused by quantization, overcoming weaknesses of the hard cropping in \cite{zhou2017fixed,RSTN}; the transformation matrix \bm{$\theta$} in RR module reasonably bridged L-Net and S-Net, which is more interpretable than dot multiplication in saliency transformation module \cite{RSTN}.

\vspace{-10pt}
\subsection{Segmentation network}
\vspace{-5pt}
RoIs with an identical scale prepared, the S-Net can perform more accurate and finer segmentation on them, since a large proportion of backgrounds containing redundant or trivial information and interference of inconsistent scales have been removed. S-Net offers great flexibility in implementation, which, like L-Net, can be any basic structure. 
Specifically, S-Net takes as input a recalibrated RoI with size of $H \times W$, the same size as that of L-Net, and outputs its predicted mask $M$ to approximate the ground-truth $M^* \in \mathbb{R}^{\mathcal{C} \times H \times W}$, where $\mathcal{C}$ denotes the category number. 
\vspace{-10pt}
\subsection{Post-processing}
\label{subsec:post-processing}
\vspace{-2pt}
All masks predicted by S-Net, they are first transformed by the inverse transformation $\theta^{-1}$ into their original sizes and locations to form their final predicted masks. Since $\theta$ is obtained in the RR, the inverse transformation $\theta^{-1}$ can be directly computed as:
\begin{equation}\vspace{-2pt}
    \theta^{-1}
    =
    \left[
    \begin{matrix}
    \frac{1}{s_x + \epsilon} & 0 & -b_x \\
    0 & \frac{1}{s_y + \epsilon} & -b_y 
    \end{matrix}
    \right].\vspace{-2pt}
\end{equation}
\noindent We add a smooth term $\epsilon=1e^{-7}$ in $\theta^{-1}$ to avoid the zero-division case.

\vspace{-10pt}
\subsection{Implementation details}
\vspace{-5pt}
\label{subsec:implementation-details}
\textbf{Loss function.}
The loss function of the proposed L-SNet can be degraded into two parts: the loss of L-Net and the loss of S-Net. 
Compared with widely adopted BCELoss, our loss is designed to adapt to the large scale variations in segmentation tasks.
Noticeably, since our proposed architecture is entirely-differentiable, the loss of S-Net can also supervise L-Net. \par

Mathematically, the training objective is to simultaneously minimize the following two loss functions defined as:
\vspace{-14pt}
\begin{align}
    L_{L-Net} &= \frac{1}{N_l} \sum_{x,y \in l}{L_{cls}{(K_{x,y}, K_{x,y}^*)}}  \notag \\ 
    & +  \frac{\lambda}{N_l^+} \sum_{x,y \in l}{\mathbbm{I}{\{K_{x,y}^* \geq 0.5\}}L_{reg}{({T}_{x,y}, {T}_{x,y}^*)}} ,\\
    L_{S-Net} &= \frac{1}{N_s} \sum_{x,y \in s}{L_{cls}{(M_{x,y}, M_{x,y}^*)}}.
    \label{loss-l-net}
\end{align}
\vspace{-14pt}

Here $L_{cls}$ is a sum of the FocalLoss \cite{focalloss} and the DiceLoss weighted by 0.2. $L_{reg}$ is the DIoULoss \cite{DIoU}. 
$N_l, N_s$ denote the sample size for L-Net and S-Net, while $N_l^+$ refers to the positive sample size for L-Net.  $\mathbbm{I}{\{\cdot\}}$ denotes an indicator function, and we set $\lambda$ as 0.5 by default. \par

\textbf{Training and inference.}
\label{subsec:training-and-inference}
At the training stage, L-Net and S-Net are tuned alternatively. 
In each iteration of L-SNet, L-Net is first tuned by minimizing $L_{L-Net}$; then the RoI is located, recalibrated by RR module, and extracted, based on a randomly selected predicted keypoint of top three highest probability and its regression result; finally, L-SNet is tuned by minimizing $L_{S-Net}$ with the extracted RoI sent to S-Net. We decay gradients back-propagated from S-Net to L-Net by a factor $\gamma=0.1$, which attenuates S-Net's indirect supervision. \par
At the inference stage, a test image is forwarded through S-Net, and the output predicted heatmap is activated (threshold by 0.5); then, centres of all activated connected components, as predicted keypoints, on that heatmap associated with their regressed distances can locate RoIs; S-Net conducts segmentation on these located RoIs and their final masks are obtained after being post-processed. \par

\vspace{-15pt}
\section{EXPERIMENTS}
\vspace{-10pt}
\subsection{Dataset and experimental settings}
\vspace{-5pt}
Experiments are conducted on TCIA Pancreas-CT dataset\cite{TCIA}, which contains 80 contrast-enhanced 3D CT scan with pancreas segmentation labelled in slice.
We take 2D image slices of 3D CT scan in the dataset as input (only use axial slices).
In practice, we use ResNet18\cite{resnet} backbones at the first stage, and ResNet34\cite{resnet} at the second stage. All backbones are pretrained on ImageNet\cite{imagenet}.
All models is optimized for 60 epochs by Adam optimizer with initial learning rate $r=1e^{-4}$. The learning rate decays by 0.1 each 25 epochs. We set $H=W=320$. The DSC and the mean IoU (mIoU) reported in Tables are the averaged score of all slices. mIoU is for the first stage, while DCS is for the second-stage. 



\begin{table}[t]
\centering\vspace{-10pt}
\caption{Performance comparison between our L-SNet and other methods on TCIA Pancreas-CT dataset with two different basic structures, FCN and U-Net.}
\vspace{-10pt}
\label{tab:mainresult}
\setlength{\tabcolsep}{1.3mm}
\begin{tabular}{p{3cm}lccc}
\hline\noalign{\smallskip}
  Method & DSC(\%) & Precision(\%) & Recall(\%)\\
\noalign{\smallskip} \hline \noalign{\smallskip}
U-Net\cite{unet}    & 86.56  & 89.60  & 90.29\\
Attention U-Net\cite{attentionunet}    & 86.88 & 89.60  & 90.29 \\
Coarse-to-fine U-Net & 87.41  &  90.05  & \textbf{91.36} \\
\textbf{L-SNet} (U-Net) &  \textbf{88.20}  & \textbf{92.50}  & 90.45 \\
\hline
FCN\cite{FCN}    & 84.28  & 88.68  & 86.85\\
Coarse-to-fine FCN & 86.16  &  91.78  & 87.84 \\
\textbf{L-SNet} (FCN) &  \textbf{87.20}  & \textbf{92.60}  & \textbf{89.07} \\
\hline
\end{tabular}\vspace{-10pt}
\end{table}

\vspace{-15pt}
\subsection{Main result}
\vspace{-5pt}
Multiple existing powerful architectures proposed by other researches, evaluated on the same dataset and with the same evaluation metrics, are compared with our L-SNet.
We implement coarse-to-fine methods \cite{zhou2017fixed, 3dc2f} in a 2D version, where two segmentation stage (coarse and fine) are connected by the RoI cropping with a 15-pixel border padding.
Two-stage models use the same backbones as L-SNet's; one-stage models use ResNet34 as backbones.
It can be observed in Table \ref{tab:mainresult} that L-SNet achieves higheset DSC regardless of different basic structures. Besides, two-stage methods are substantially accurate than one-stage methods. Importantly, L-SNets with FCNs and U-Nets still outperform their coarse-to-fine counterparts by 1.04\% and 0.79\% DSC, respectively. These all prove the effectiveness of our L-SNet architecture. \par


\begin{table}[ht]\vspace{-8pt}
\centering
\caption{The effectiveness of different components of our method. 
}\vspace{-10pt}
\label{tab:ablation}
\setlength{\tabcolsep}{2mm}
\begin{tabular}{cccccc}  
\hline\noalign{\smallskip}
  S-Net & Loss & L-Net & RR & DSC(\%)\\
\hline
\checkmark & & & & 84.69\\
\checkmark  & \checkmark &  &  & 86.51\\
\checkmark  & \checkmark & \checkmark & & 87.86\\
\checkmark  & \checkmark & \checkmark  & \checkmark & \textbf{88.20}\\
\hline
\end{tabular}\vspace{-10pt}
\end{table}
\vspace{-15pt}
\subsection{Ablation study}
\vspace{-5pt}
We decompose the accomplishment of L-SNet into four dominant factors (see Table \ref{tab:ablation}) which are studied in ablation: S-Net, our loss function versus BCELoss, L-Net, and RR operation versus cropping. All experiments are performed on U-Net basic structure.


\vspace{-10pt}
\subsection{Detection v.s. segmentation at the first stage}
\vspace{-5pt}
As is mentioned in Section \ref{sec:introduction}, in the coarse-to-fine method, the segmentation model at the first stage actually executing the detection task is not so efficient as the detection model. We conduct an experiment where U-Net is replaced by our L-Net at the first stage of the coarse-to-fine method \cite{zhou2017fixed,3dc2f}. As Table \ref{tab:landu} shows, both mIoU and final DCS are significantly improved when L-Net is utilized with only a negligible computation increase. 
\begin{table}[ht]
\centering \vspace{-5pt}
\caption{Comparison between L-Net and U-Net as the first-stage network in the coarse-to-fine method.} 
\vspace{-8pt}
\label{tab:landu}
\setlength{\tabcolsep}{2mm}
\begin{tabular}{cccc}  
\hline\noalign{\smallskip}
First stage    & mIoU(\%) & DCS(\%) & Params(M)\\ 
\hline
U-Net & 83.83 &  87.41 & \textbf{14.33}\\
L-Net & \textbf{86.58} &  \textbf{87.86} & 14.50\\
\hline
\end{tabular}
\end{table}
\vspace{-20pt}
\subsection{The effect of decay factor }
\vspace{-5pt}
The decay factor $\gamma$ reflects the intensity of S-Net's supervision on L-Net. 
Table \ref{tab:gil} provides the comparison results when using U-Net as the basic structure. 
Although a larger $\gamma$ may degenerate the detection performance, it leads to a better final segmentation score and a smaller gap between the mIoU of training and validation. 
It also can be seen that though $\gamma=1$ achieves the smallest gap between training and validation, the final DSC is low. This phenomenon shows that the RR module works in a more complex way, not just to reduce the gap between training and validation. \par
\begin{table}[ht]
\centering \vspace{-8pt}
\caption{The effect of the decay factor $\gamma$ on detection (mIoU) and segmentation (DSC) results. 
}
\label{tab:gil}\vspace{-10pt}
\setlength{\tabcolsep}{2mm}
\begin{tabular}{ccccc}  
\hline\noalign{\smallskip}
{$\gamma$}    & 0 & 0.01 & 0.1 & 1\\
\hline
mIoU in training & \textbf{92.03} & 88.09 & 88.24 & 86.84\\
mIoU in validation & \textbf{86.58} & 84.88 & 84.98 & 84.68 \\
DSC(\%) & 87.86 & 88.05 & \textbf{88.20} & 87.21 \\
\hline
\end{tabular}
\end{table}


\vspace{-24pt}
\section{Conclusion}
\vspace{-8pt}


In this paper, we analysed previous works on medical image segmentation and proposed a new architecture, L-SNet. In L-SNet, segmentation task undertaken by both L-Net and S-Net: L-Net is designed for localisation and S-Net is designed for segmentation. RR module connected L-Net and S-Net and establish the overall differentiability. Experiments show that every module in L-SNet, concluding L-Net, S-Net, and RR module, improved the final DSC. Conclusively, with all modules in L-SNet, we outperform coarse-to-fine methods consistently.

\vspace{-15pt}
\section{Acknowledgments}
\label{sec:majhead}\vspace{-8pt}
This work was supported by the General Program of National Natural Science Foundation of China (NSFC) (Grant No. 61806147), and Shanghai Natural Science Foundation of China (Grant No. 18ZR1441200).

\bibliographystyle{IEEEbib}
\bibliography{strings,refs}

\end{document}


%
\maketitle
\begin{figure*}[hb]
    \centering
    \resizebox{\textwidth}{!}{
    \includegraphics{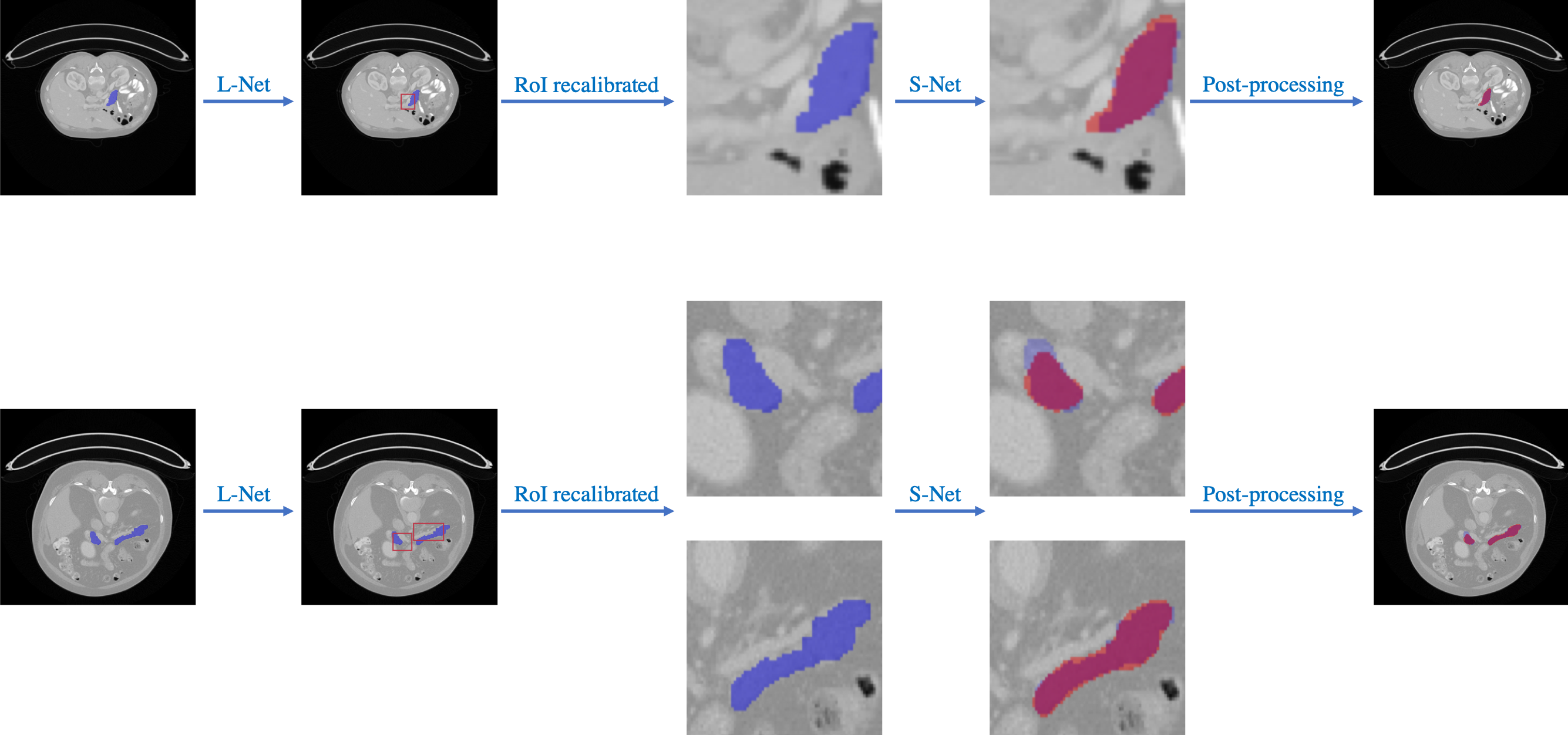}
    }
    \caption{
    Visualization of L-SNet's workflow to perform medical image segmentation, which is illustrated by two samples. The ground truth areas are shown in purple; the predicted areas are shown in pink. RoIs located by L-Net and padded with extra pixels are bounded by red boxes.}
    \label{fig:s1}
\end{figure*}

To illustrate the L-SNet's prediction process vividly, we take two samples (a trivial one and a complex one), pass them through L-SNet, and present their intermediate results (see Fig.\ref{fig:s1}).

\begin{figure*}[ht]
    \centering
    \resizebox{\textwidth}{!}{
    \includegraphics{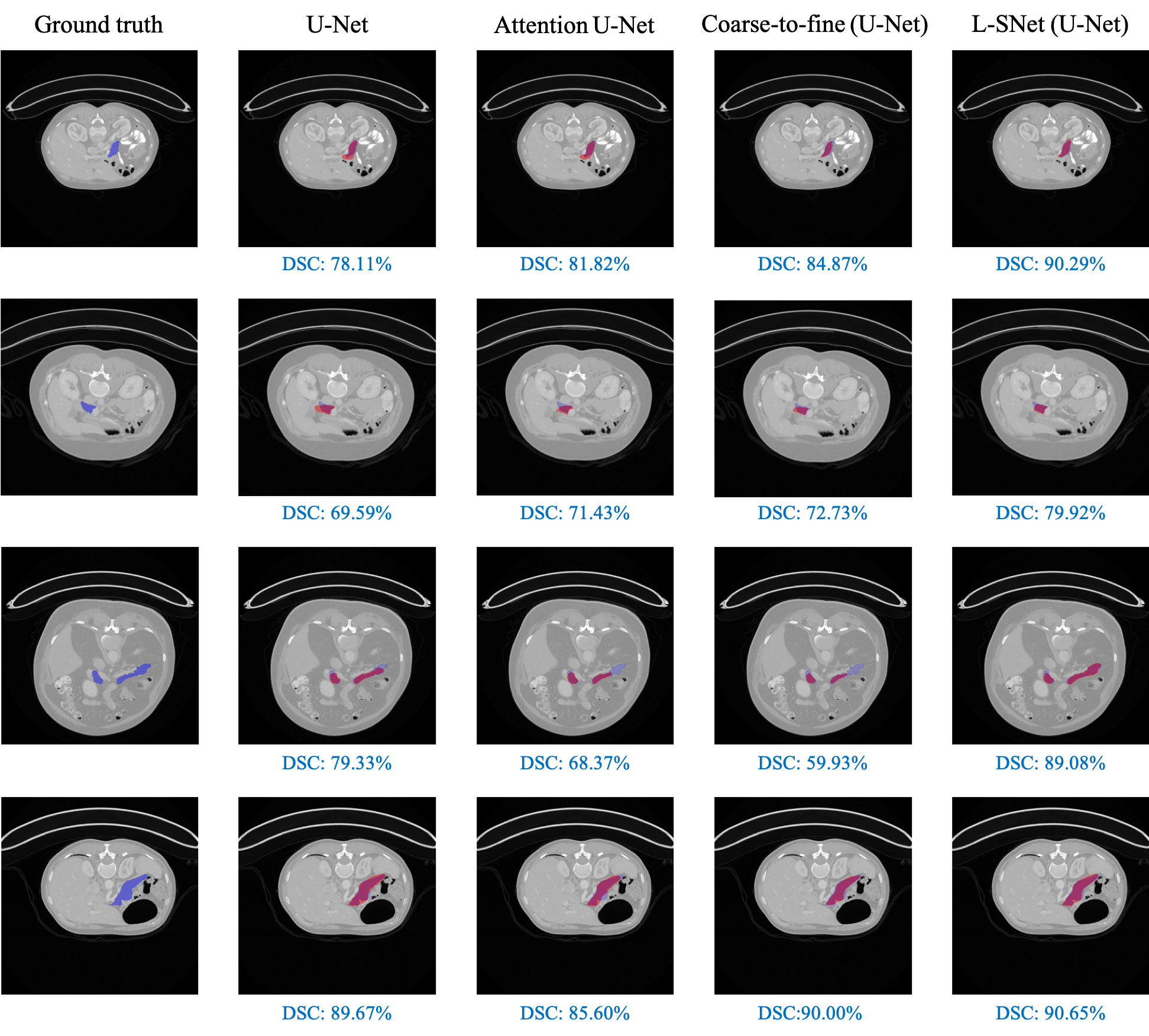}
    }
    \caption{
    Segmentation results of four samples are compared among our L-SNet and other three proposed models. The first two samples are with small scale RoIs; the third sample is with multiple RoIs; the last sample has an RoI with a larger scale. Each prediction is displayed with its segmentation result and with DSC score located below (in percentage).}
    \label{fig:s2}
\end{figure*}


Fig.\ref{fig:s2} displays the segmentation performance comparison of four samples among different models.